\pdfoutput=1

\documentclass[11pt]{article}
\usepackage{authblk}
\usepackage[hyperref]{emnlp2021}

\usepackage{times}
\usepackage{latexsym}
\usepackage{url}

\usepackage{enumitem}
\usepackage{graphicx}
\usepackage{multirow}
\usepackage{caption}
\usepackage{subcaption}

\usepackage[T1]{fontenc}

\usepackage[utf8]{inputenc}

\usepackage{microtype}
\title{GradTS: A Gradient-Based Automatic Auxiliary Task Selection Method \\Based on Transformer Networks}

\author[1]{Weicheng Ma}
\author[2$\dagger$]{Renze Lou}
\author[3$\dagger$]{Kai Zhang}
\author[1]{Lili Wang}
\author[4]{Soroush Vosoughi}
\affil[1,4]{Department of Computer Science, Dartmouth College} 
\affil[2]{Department of Computer Science, Zhejiang University City College}
\affil[3]{Department of Computer Science and Technology, Tsinghua University}
\affil[1]{\texttt{\{first.last\}.gr@dartmouth.edu}}
\affil[2]{\texttt{marionojump0722@gmail.com}}
\affil[3]{\texttt{drogozhang@gmail.com}}
\affil[4]{\texttt{soroush.vosoughi@dartmouth.edu}}

\date{}

\begin{document}
\maketitle
\renewcommand{\thefootnote}{\fnsymbol{footnote}}
\footnotetext[2]{Work done when interning at the Minds, Machines, and Society Lab at Dartmouth College.}
\renewcommand{\thefootnote}{\arabic{footnote}}
\begin{abstract}
A key problem in multi-task learning (MTL) research is how to select high-quality auxiliary tasks automatically.
This paper presents GradTS, an automatic auxiliary task selection method based on gradient calculation in Transformer-based models.
Compared to AUTOSEM, a strong baseline method, GradTS improves the performance of MT-DNN with a bert-base-cased backend model, from 0.33\% to 17.93\% on 8 natural language understanding (NLU) tasks in the GLUE benchmarks.
GradTS is also time-saving since (1) its gradient calculations are based on single-task experiments and (2) the gradients are re-used without additional experiments when the candidate task set changes.
On the 8 GLUE classification tasks, for example, GradTS costs on average 21.32\% less time than AUTOSEM with comparable GPU consumption.
Further, we show the robustness of GradTS across various task settings and model selections, e.g. mixed objectives among candidate tasks.
The efficiency and efficacy of GradTS in these case studies illustrate its general applicability in MTL research without requiring manual task filtering or costly parameter tuning.
\end{abstract}

\section{Introduction}
MTL \cite{mtl-orig} is widely used in NLU research to improve the performance of machine learning (ML) models by enlarging the training data size with datapoints related to the primary tasks.
However, its efficacy is largely affected by the selection of auxiliary tasks.
The auxiliary task selection problem is addressed mainly under two settings.
The first setting treats each task as a whole.
For example, \citet{select-1} assess task relatedness by exhaustive experiments in all task pairs.
Nonetheless, high pairwise task correlations are often not decisive features for choosing auxiliary tasks.
\citet{select-2} train a policy for task selection through counterfactual estimation, but their learned policy brings improvements only to one out of nine tasks on GLUE benchmarks \cite{glue-orig}.
The second setting subsamples training instances from auxiliary tasks, e.g. with Bayesian optimization \cite{subsampling-1}, but these methods are time- and resource-consuming due to their reliance on multi-task experiments involving all the candidate tasks.
AUTOSEM \cite{autosem-orig} combines the two settings into one method, selecting candidate tasks with Thompson sampling and deciding the ratio with which to draw training instances from the selected tasks via a Gaussian Process.
Despite the higher quality of the auxiliary task sets it generates, AUTOSEM is still costly, similar to \citet{subsampling-1}.

To design a better-performing and less costly auxiliary task selection method, we take advantage of the characteristics of Transformer networks \cite{transformer-orig}.
Prior research reveals that in a Transformer-based model, each attention head attends on specialized linguistic features \cite{probe-1,probe-2,probe-3,probe-4,probe-5,probe-6}.
Since important linguistic features strongly correlate with the goals of tasks, we further hypothesize that a good auxiliary task shares key linguistic features with the primary task.
Thus, we address the auxiliary task selection problem by maximizing the overlap of important heads in a Transformer-based model between primary and auxiliary tasks.
As \citet{head-importance-1} claim, the importance of attention heads to a task can be approximated by the absolute gradients accumulated at each head.
We design our auxiliary task selection method, GradTS, accordingly, by ranking the importance of attention heads for each individual task and modeling the correlation between each pair of tasks with their head rankings.
By greedily selecting the tasks most closely related to the primary task, GradTS constructs auxiliary task sets through trial experiments (GradTS-trial).
GradTS also enables task subsampling to further optimize auxiliary task sets.
To achieve this goal, we design another setting of GradTS (GradTS-fg) that first assesses the correlations between the primary task and each training instance in an auxiliary task selected by GradTS-trial and then filters the training instances via thresholding.

We assess the strength of GradTS via MTL evaluations 
on 8 GLUE classification tasks.
We use AUTOSEM and AUTOSEM-p1 \footnote{We refer to the AUTOSEM method without task subsampling as AUTOSEM-p1.} as our baselines since AUTOSEM is among the most advanced auxiliary task selection methods in the NLP field and because it features both task selection and task subsampling.
For consistency, we use the bert-base-cased model as the backend model of GradTS, AUTOSEM, and the MTL framework.
Results show that GradTS-trial produces better auxiliary task sets than AUTOSEM-p1 in all 8 GLUE tasks while costing on average 6.73\% less time.
In experiments with task subsampling, GradTS-fg again shows superior strength to AUTOSEM on all 8 tasks while costing 21.32\% less time.
These results strongly support the efficacy and efficiency of GradTS.

In addition to the main experiments, we compare GradTS to multiple intuitive auxiliary task selections to show its high performance.
We also conduct case studies to show that GradTS is effective and robust on difficult tasks or candidate tasks with mixed objectives.
These findings reflect the general applicability of GradTS in various task settings.
In comparison, auxiliary task sets produced by AUTOSEM and AUTOSEM-p1 are often not optimal in these complicated scenes.
Further, GradTS reuses the head rankings when the candidate task set grows larger, which makes it even more time- and resource-efficient than existing methods.

The contributions of this paper are three-fold:
\begin{itemize}[leftmargin=*,topsep=0pt]
\setlength{\itemsep}{0cm}
\setlength{\parskip}{0cm}
    \item we propose GradTS, an automatic auxiliary task 
 selection method based on gradient calculation in pre-trained Transformer-based models;
    \item we illustrate the efficacy and efficiency of GradTS through comprehensive MTL evaluations; and
    \item we show, through case studies, the superior capability and robustness of GradTS to complicated candidate task settings compared to both AUTOSEM and auxiliary task selections based on human intuition.
\end{itemize}

\section{Datasets}
\begin{table}[h]
\begin{tabular}{|c|c|c|c|}
\hline
Datasets    & OBJ & LBL & Training Size \\ \hline
CoLA        & CLS & 2   & 8,550          \\ \hline
MRPC        & CLS & 2   & 3,667          \\ \hline
MNLI        & CLS & 3   & 392,701          \\ \hline
QNLI        & CLS & 2   & 104,742          \\ \hline
QQP         & CLS & 2   & 363,845          \\ \hline
RTE         & CLS & 2   & 2,489          \\ \hline
SST-2       & CLS & 2   & 67,348           \\ \hline
WNLI        & CLS & 2   & 634           \\ \hline
STSB        & RGR & -   & 5,748            \\ \hline
POS         & SL  & -   &14,040\\ \hline
NER         & SL  & -   &14,987\\ \hline
SC    & SL  & -   &8,936\\ \hline
MELD        & CLS & 7   &9,988\\ \hline
Dyadic-MELD & CLS & 7   &12,839\\ \hline
\end{tabular}
\caption{Details of datasets used in this paper. OBJ denotes task objectives and LBL refers to the number of classes for classification tasks. Training size represents the number of training instances in each task. CLS, RGR, and SL are classification, regression, and sequence labeling tasks, respectively.}
\label{tbl:dataset-details}
\end{table}
Following \citet{autosem-orig}, we use the 8 classification tasks in GLUE benchmarks \cite{glue-orig}, namely CoLA, MRPC, MNLI, QNLI, QQP, RTE, SST-2, and WNLI, in our main experiments.
We apply the standard split of these datasets as \citet{glue-orig} describe. \footnote{We report scores on the development set of GLUE tasks due to the submission quota limit. We sample 10\% training instances of each GLUE task with a random seed of 42 for choosing thresholds and selecting auxiliary tasks.}

We also use one regression and three sequence labeling tasks in our case studies about the efficacy of GradTS on candidate tasks with mixed training objectives.
These tasks include STSB from GLUE benchmarks, Part-of-Speech tagging (POS) from Universal Dependencies \footnote{https://universaldependencies.org/}, Named Entity Recognition (NER) from CoNLL-2003 challenges \cite{ner-orig}, and Syntactic Chunking (SC) from CoNLL-2000 shared tasks \cite{chunking-orig}.
The official data split of all these datasets is applied.

Additionally, we introduce MELD and Dyadic-MELD datasets \cite{meld-orig} to verify the applicability of GradTS to tasks that are difficult for its backend model.
While these two tasks are multi-modal emotion recognition tasks, we use only the textual data in the experiments.
The MELD and Dyadic-MELD datasets are annotated with 7 emotion labels.
The bert-base-cased model achieves F-1 scores less than $50\%$ on both tasks, lower than its performance on most GLUE classification tasks.

Details of the datasets are displayed in Table \ref{tbl:dataset-details}.
We evaluate both accuracy and F-1 scores for MRPC and QQP, accuracy for QNLI, RTE, SST-2, MNLI \footnote{We report accuracy scores separately on the matched and mismatched splits of MNLI.} and WNLI, Matthew's correlation coefficient (MCC) for CoLA, Pearson's correlation coefficient and Spearman's correlation coefficient for STSB, and F-1 score for POS, NER, SC, MELD, and Dyadic-MELD tasks.

\section{Methodology}
We design GradTS based on the hypothesis that better auxiliary tasks share more important linguistic features with the primary task.
Since each attention head in a Transformer-based model functions similarly as a standalone feature extractor on a specialized set of features, we approximate the important feature set of each task by the heads contributing the most to the task.
As the key feature sets are task-specific, GradTS does not require multi-task experiments to rank auxiliary tasks given a primary task.
This makes GradTS a time- and resource-economic method especially when the set of candidate auxiliary tasks is large or growing.

GradTS consists of three successive modules responsible for (1) ranking attention heads for a task based on their contributions, (2) ranking auxiliary tasks based on inter-task correlations, and (3) finalizing the auxiliary task sets, respectively.

\subsection{Attention Head Ranking Module}
We estimate the importance of attention heads to a task using the absolute gradients accumulated at each head, following \citet{head-importance-1}.
Specifically, we achieve the goal in four steps:
(1) We fine-tune a pre-trained Transformer-based model on a task.
(2) We repeat the fine-tuning step on the training set of the task with the fine-tuned model, without updating parameters, to get gradients of the model.
(3) We sum up the absolute gradients accumulated at each attention head during the last fine-tuning step.
(4) We layer-wise normalize the accumulated gradients and scale the gradients to the range [0, 1] globally to represent the importance of each head for the given task.

In practice, we use pre-trained bert-base-cased model as the backend of GradTS and we fine-tune the model for three epochs before starting to accumulate gradients on each head \footnote{Our preliminary experiments show that fine-tuning the backend model for three to seven epochs at the warm up stage does not have much effect on the predictions of GradTS.}.
This fine-tuning stage is designed to avoid large gradients on unimportant heads when the model is exposed to a downstream task for the first time.

\subsection{Auxiliary Task Ranking Module}
Given a primary task, we rank each candidate auxiliary task by the correlation between its head ranking matrix and that of the primary task.
As \citet{tau-vs-pho} suggest, we use Kendall's rank correlation coefficients (Kendall's $\tau$) since the importance scores of heads seldom result in a tie, based on our observations.
We visualize the head importance matrix of the bert-base-cased model on MRPC and task correlations for the 8 GLUE classification tasks in Figures \ref{fig:head-ranking-example} and \ref{fig:task-corr} in Appendix.

While the rankings of auxiliary tasks produced by GradTS are intuitive in some cases, e.g. the three natural language inference (NLI) tasks are good auxiliary tasks for each other, the correlation scores between many seemingly unrelated tasks, e.g. WNLI and CoLA, are also high.
This reveals the difficulty of manually designing auxiliary task sets since the factors affecting the appropriateness of auxiliary tasks are multi-faceted, e.g. text lengths and label distributions.
As a result, designing automatic methods for selecting auxiliary tasks makes up a crucial part of MTL research, especially at a time when candidate auxiliary tasks are rapidly growing both larger in amount and more complex.

\subsection{Auxiliary Task Selection Module}
After obtaining the rankings of candidate auxiliary tasks for each primary task, we finalize the auxiliary task selection process through trial experiments.
We also study the potential of GradTS to subsample the selected auxiliary tasks.
Our experiments show that with one additional fine-tuning pass of its backend model on the individual tasks, GradTS produces subsampled auxiliary training sets higher in quality than the task-level selections.

We introduce the two settings of GradTS to select tasks from the task correlations as follows:\\
\textbf{[Task-level Trial-based]}
We select auxiliary tasks greedily under this setting.
Starting from the most closely-correlated task to a primary task, we keep adding tasks to the auxiliary task set and run MTL evaluations on the primary task and all the chosen auxiliary tasks.
GradTS stops adding new tasks when the evaluation score starts to decrease on the validation set we leave for parameter tuning and finalizes the auxiliary task set with the tasks chosen at the previous step.\\
\textbf{[Instance-level]}
We re-run the base model of GradTS on all the individual tasks once, with gradient calculation but not parameter updates.
For each instance, we take the absolute value of its gradients on all the attention heads, layer-wise normalize the gradients, and scale the numbers to the range [0, 1].
Then we calculate and record the correlation score between the normalized gradient matrix and the head ranking matrix of each candidate auxiliary task.
Last, we use a threshold to select auxiliary training instances from tasks chosen by the task-level trial-based method to form a subsampled auxiliary task set.
The threshold we use in this paper is tuned by experiments on RTE, MRPC, and CoLA tasks, which is a Kendall's $\tau$ of 0.42.

We refer to GradTS with the task-level trial-based and instance-level task selection settings as GradTS-trial and GradTS-fg, respectively.

\section{Experiments and Analysis} \label{sct:experiments}
\begin{figure}[t]
    \centering
    \begin{subfigure}[b]{.48\linewidth}
    \centering
        \includegraphics[height=.6\linewidth]{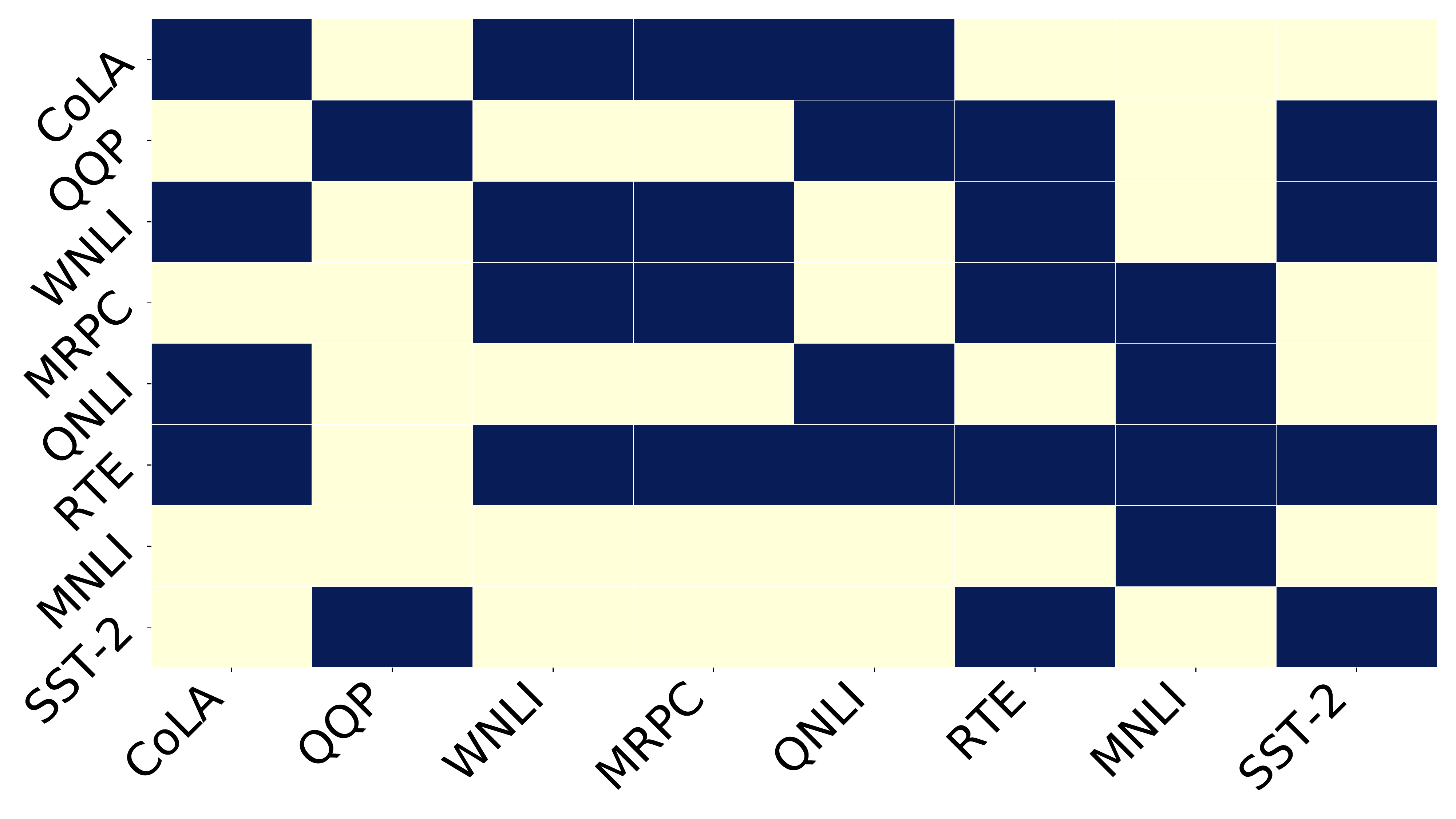}
        \caption{GradTS-trial}
    \end{subfigure}
    ~
    \begin{subfigure}[b]{.48\linewidth}
    \centering
        \includegraphics[height=.6\linewidth]{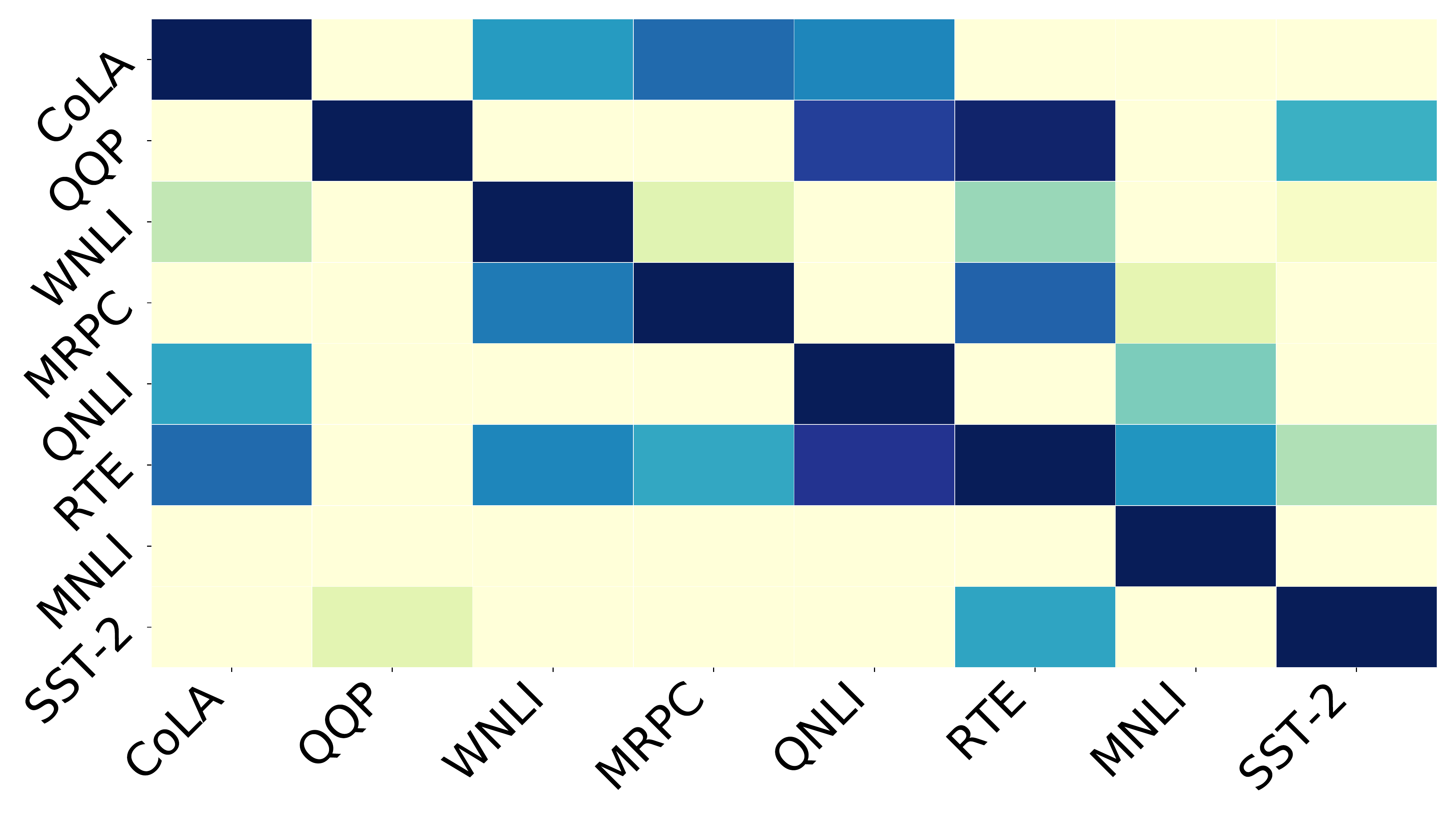}
        \caption{GradTS-fg}
    \end{subfigure}
    
    \begin{subfigure}[b]{.48\linewidth}
    \centering
        \includegraphics[height=.6\linewidth]{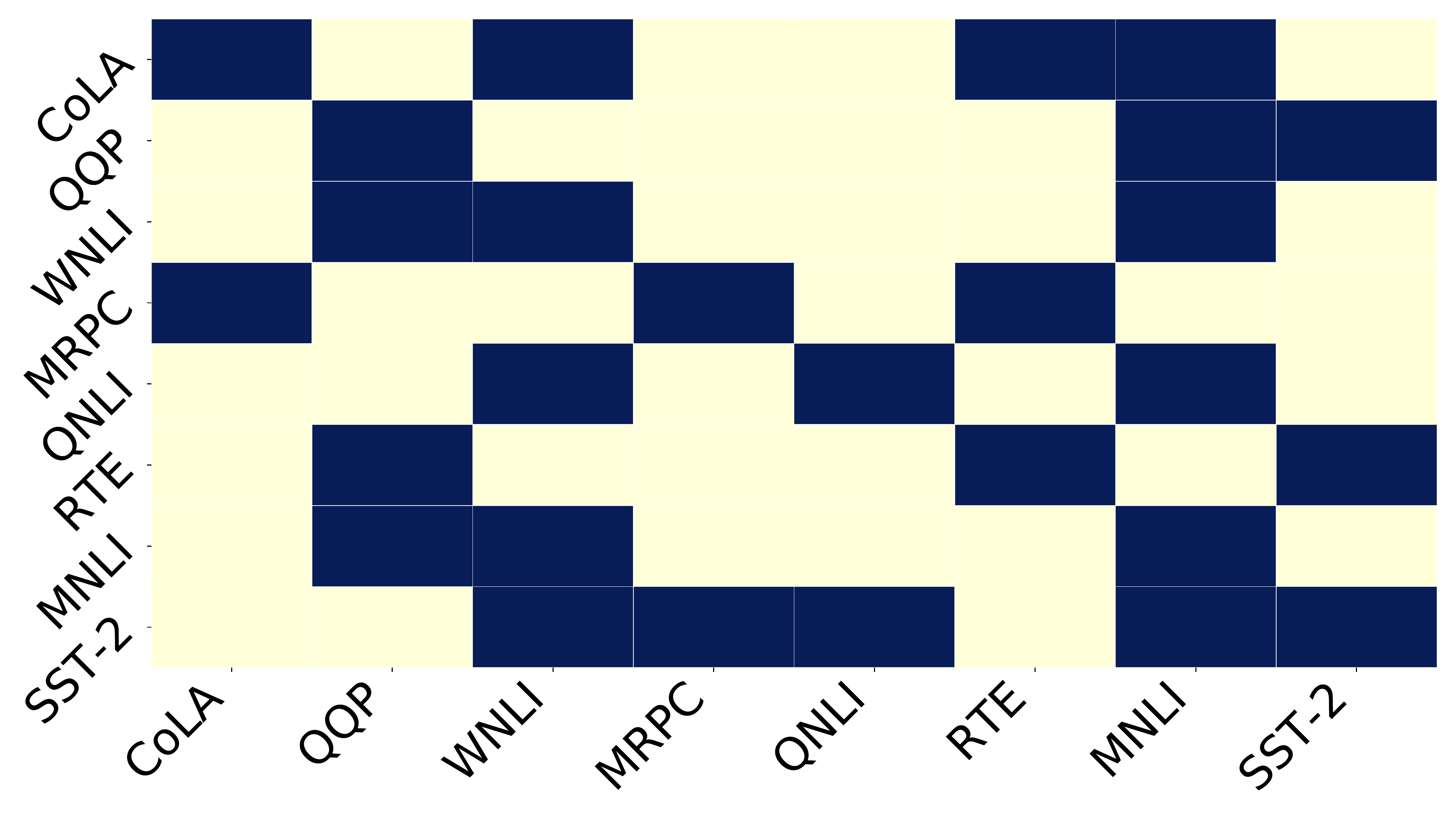}
        \caption{AUTOSEM-p1}
    \end{subfigure}
    ~
    \begin{subfigure}[b]{.48\linewidth}
    \centering
        \includegraphics[height=.6\linewidth]{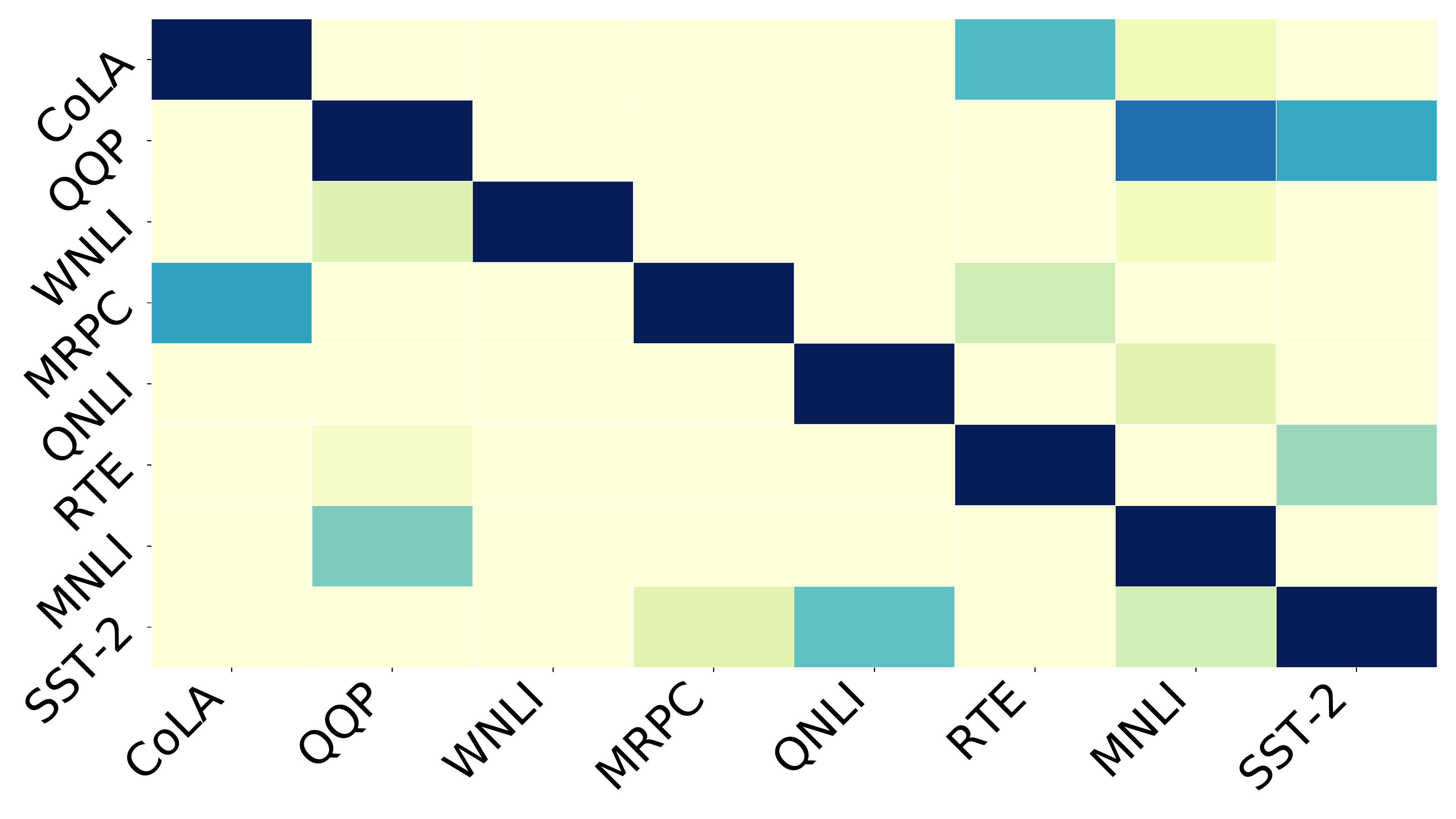}
        \caption{AUTOSEM}
    \end{subfigure}
    \caption{Task selection results by two AUTOSEM and two GradTS methods on 8 GLUE classification tasks. Y and X axes represent primary and auxiliary tasks, respectively. Darker color in a cell indicates that a larger portion of an auxiliary task is selected.}
    \label{fig:aux-task-rankings}
\end{figure}
\begin{table*}[t]
\centering
\begin{tabular}{|l|c|c|c|c|c|c|c|c|}
\hline
\multicolumn{1}{|c|}{\multirow{2}{*}{Methods}} & \multicolumn{8}{c|}{Primary Tasks}                                                                                                                      \\ \cline{2-9} 
\multicolumn{1}{|c|}{}                         & CoLA           & MRPC                 & MNLI                 & QNLI           & QQP                  & RTE            & SST-2          & WNLI           \\ \hline
Single-Task                                    & 51.00          & 78.19/84.58          & 83.52/83.88          & 90.26          & 90.34/87.15          & 63.18          & 91.63          & 53.52          \\ \hline
NO-SEL                                         & 50.89          & 79.17/84.96          & 82.68/83.18          & 90.06          & 90.44/87.16          & 66.98          & 91.40          & 47.89          \\ \hline
AUTOSEM-p1                                     & 54.13          & 81.62/86.77          & 83.41/83.40           & 90.48          & 90.65/87.39          & 67.15          & 92.09          & 48.97          \\ \hline
AUTOSEM                                        & 56.25          & 84.24/75.25          & 83.65/83.43          & 90.13          & 90.67/87.44          & 67.87          & 91.74          & 49.29          \\ \hline
GradTS-trial                                   & 55.24          & 83.58/88.35          & 83.95/83.55          & 90.62          & 90.87/87.72          & 76.53          & 92.31          & 54.93          \\ \hline
GradTS-fg                                      & \textbf{58.38} & \textbf{84.07/88.74} & \textbf{83.79/83.96} & \textbf{90.87} & \textbf{90.89/87.73} & \textbf{76.90} & \textbf{92.63} & \textbf{57.75} \\ \hline
\end{tabular}
\caption{MTL evaluation results on 8 GLUE classification tasks. Single-Task refers to the single-task performance of the bert-base-cased model. NO-SEL includes all the candidate tasks in the auxiliary task set of each primary task. The highest score for each task is in bold.}
\label{tbl:main-experiment-results}
\end{table*}
\begin{table}[]
\centering
\begin{tabular}{|l|r|r|}
\hline
\multicolumn{1}{|c|}{Methods} & \multicolumn{1}{c|}{Time Cost} & \multicolumn{1}{c|}{GPU Usage} \\ \hline
AUTOSEM-p1                    &114&37,003\\ \hline
AUTOSEM                       &194&46,361\\ \hline
GradTS-trial                  &107&39,551\\ \hline
GradTS-fg                     &153&35,610\\ \hline
\end{tabular}
\caption{Average time and GPU consumption for 4 auxiliary task selection methods on each of the 8 GLUE classification tasks. The units are minutes for time cost and megabytes for GPU usage.}
\label{tbl:runtime}
\end{table}
\begin{table*}[t]
\centering
\begin{tabular}{|l|r|r|r|r|r|r|}
\hline
\multicolumn{1}{|c|}{\multirow{2}{*}{\begin{tabular}[c]{@{}c@{}}Primary \\ Tasks\end{tabular}}} & \multicolumn{6}{c|}{Methods}                                                                                                                                                                         \\ \cline{2-7} 
\multicolumn{1}{|c|}{}                                                                          & \multicolumn{1}{c|}{AUTOSEM-p1} & \multicolumn{1}{c|}{AUTOSEM} & \multicolumn{1}{c|}{GradTS-trial} & \multicolumn{1}{c|}{GradTS-fg} & \multicolumn{1}{c|}{Single-Task} & \multicolumn{1}{c|}{NO-SEL} \\ \hline
CoLA                                                                                            & 57.04                           & 57.38                        & 61.81                             & \textbf{62.66}                 & 51.00                            & 46.50                        \\ \hline
MRPC                                                                                            & 78.43/85.14                     & 80.39/85.97                  & 82.35/87.32                       & \textbf{83.33/88.40}           & 78.19/84.58                      & 78.92/84.91                 \\ \hline
MNLI                                                                                            & 81.18/81.70                      & 83.67/83.56                  & 83.53/83.42                       & \textbf{83.51/84.06}           & 83.52/83.88                      & 83.24/83.34                 \\ \hline
QNLI                                                                                            & 90.63                           & 90.57                        & 90.66                             & \textbf{90.85}                 & 90.26                            & 90.52                       \\ \hline
QQP                                                                                             & 90.89/87.32                     & 90.36/87.21                  & 90.65/87.39                       & \textbf{90.72/87.47}           & 90.34/87.15                      & 89.37/85.72                 \\ \hline
RTE                                                                                             & 75.45                           & 76.73                        & 75.45                             & \textbf{77.62}                 & 63.18                            & 72.20                        \\ \hline
SST-2                                                                                           & 91.86                           & 92.55                        & 91.86                             & \textbf{92.66}                 & 91.63                            & 89.30                        \\ \hline
WNLI                                                                                            & 45.07                           & 52.11                        & 56.34                             & \textbf{57.75}                 & 53.52                            & 43.66                       \\ \hline
MELD                                                                                            & 39.96                           & 42.59                        & 45.36                             & \textbf{47.02}                 & 39.14                            & 39.26                       \\ \hline
D-MELD                                                                                          & 43.46                           & 43.37                        & 47.53                             & \textbf{47.61}                 & 37.44                            & 37.44                       \\ \hline
\end{tabular}
\caption{MTL evaluation results with AUTOSEM and GradTS auxiliary task selection methods on 10 classification tasks. Single-Task indicates single-task performance of bert-base-cased and NO-SEL indicates performance of MT-DNN, with the bert-base-cased backend, trained on all 10 tasks. D-MELD refers to Dyadic-MELD.}
\label{tbl:res-cls-with-meld}
\end{table*}
\subsection{Experimental Settings}
To show the strength of GradTS, we run evaluations with MT-DNN \cite{mt-dnn-orig} as the MTL evaluation framework on 8 classification tasks in GLUE benchmarks.
The bert-base-cased model is used as the backend of MT-DNN and all the auxiliary task selection methods.
For tasks whose input contains multiple sentences, we concatenate the sentences together with a [SEP] token in between.
We use the Huggingface \cite{huggingface-orig} implementation of BERT \cite{bert-orig} and other pre-trained models in this paper.
In each experiment, we fine-tune MT-DNN for 7 epochs with a learning rate of 5e-5 and report the highest score \footnote{We apply the same set of hyper-parameters in all the experiments for fair comparison. We also use official dataset splits to minimize randomness in all  our experiments.}.

\subsection{Auxiliary Task Selection Results}
Figure \ref{fig:aux-task-rankings} shows the auxiliary task sets selected by AUTOSEM-p1, AUTOSEM, GradTS-trial, and GradTS-fg methods.
Each auxiliary task is labeled as 1 (selected) or 0 (not selected) for methods under the task-level auxiliary task selection setting (AUTOSEM-p1 and GradTS-trial).
The percentage of selected training data amount in each auxiliary task is reflected for AUTOSEM and GradTS-fg.

While some common task combinations appear in the auxiliary task sets constructed by both GradTS-trial and AUTOSEM-p1, e.g. CoLA-WNLI and QNLI-MNLI, the two methods generally make very different selections.
We note that GradTS-trial usually generates larger auxiliary task sets than AUTOSEM-p1 on tasks with small training data size, e.g. WNLI, RTE, and MRPC.
Different from AUTOSEM-p1 which balances exploitation with exploration at the task selection phase, the auxiliary task ranking mechanism of GradTS-trial is in full charge of controlling the risk of selecting improper auxiliary tasks.
The task selection module of GradTS-trial greedily chooses auxiliary tasks based on the task rankings and it is thus more likely to also select auxiliary tasks marginally improving the performance of the primary task than AUTOSEM-p1.
There are more disagreements between the task selection ratios of GradTS-fg and AUTOSEM than the task-level selections.
For example, while WNLI is constantly discarded by AUTOSEM at its second phase due to the small size of WNLI, GradTS-fg ranks WNLI highly for three primary tasks (CoLA, MRPC, and RTE).
Benefiting from its training instance ranking mechanism which treats each record independently, GradTS-fg is robust to the higher overall impact of a few noisy instances in smaller datasets.
As such, GradTS has a lower chance of underestimating the importance of small auxiliary datasets than AUTOSEM.

While some auxiliary task selection results are intuitive, they are mostly beyond the scope of manual designs.
For example, QQP is not chosen by either AUTOSEM or GradTS as a good auxiliary task for CoLA or MRPC, despite its large size.
It is also counter-intuitive that GradTS does not select MNLI or QNLI into the auxiliary task set of WNLI though these tasks share similar goals.
Due to the gap between the automatic auxiliary task selection results and human intuitions, we assess the strength of these task selection methods via MTL evaluations and show the results in Table \ref{tbl:main-experiment-results}.

\subsection{MTL Evaluation Results}
While MTL is designed to enhance model performance, our evaluations reveal that simply using all the available auxiliary tasks without selection is not sufficient.
Despite the enlarged training dataset, MTL with all the candidate auxiliary tasks brings only marginal improvements to 3 out of the 8 GLUE classification tasks.
On the contrary, the MTL performance is generally higher than single-task evaluation scores when an auxiliary task selection method is applied.
We attribute this phenomenon to the greater discrepancies in some primary-auxiliary task combinations without carefully selecting auxiliary tasks.
These results show that while MTL provides a promising way to boost the performance of ML models, a good automatic auxiliary task selection method is necessary.

Between the two task-level auxiliary task selection methods, GradTS-trial produces better auxiliary task sets than AUTOSEM-p1 for all the 8 primary tasks.
MTL performance with GradTS-trial also beats the single-task baseline in all the evaluations, while AUTOSEM-p1 produces low-quality auxiliary task sets on tasks whose training sets are extremely large (MNLI) or small (WNLI) compared to the other tasks.
This demonstrates that GradTS-trial is more robust to the design of candidate auxiliary task sets than AUTOSEM-p1. 
Though the auxiliary task sets selected by AUTOSEM-p1 and GradTS-trial overlap a lot for CoLA, MRPC and RTE, no training instance is drawn from WNLI in its second phase, resulting in large performance gaps between AUTOSEM and GradTS-fg on these tasks.
For comparison, GradTS-fg samples 59.94\%, 70.98\%, and 60.25\% of the WNLI dataset, respectively, for CoLA, MRPC, and RTE, and achieves 3.79\%, 17.93\%, and 13.30\% higher MTL evaluation scores than AUTOSEM on these tasks.
Despite the generally higher fragility of small datasets to noisy annotations, these datasets may contain useful datapoints as auxiliary training instances and should not be completely ignored.
GradTS-fg subsamples tasks on the instance level, which is more efficient and flexible in picking highly-correlated training instances than the second phase of AUTOSEM.

\subsection{Running Time Analysis}

\begin{figure}[t]
    \centering
    \begin{subfigure}[b]{.48\linewidth}
    \centering
        \includegraphics[width=1\linewidth]{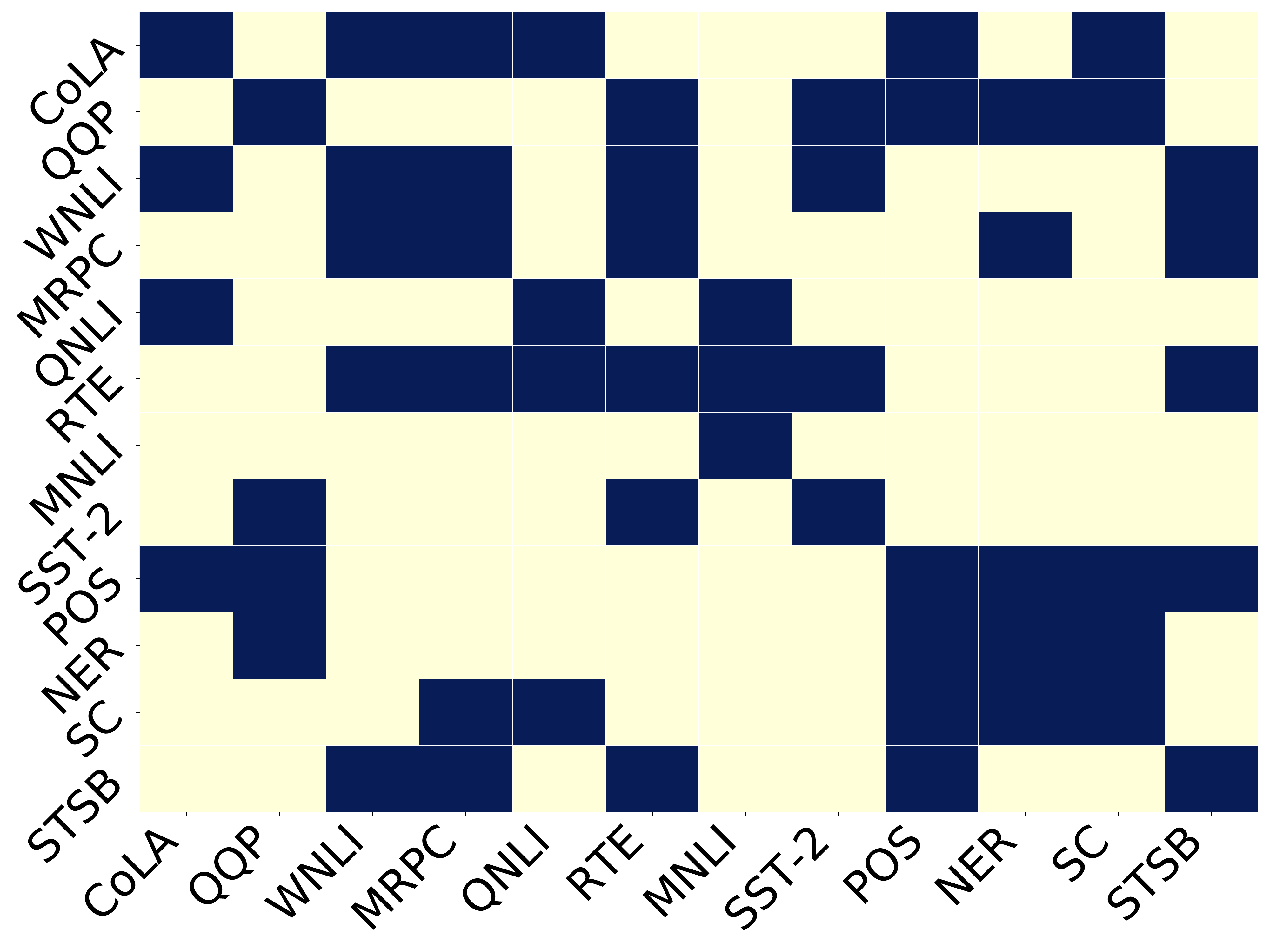}
        \caption{GradTS-trial}
    \end{subfigure}
    ~
    \begin{subfigure}[b]{.48\linewidth}
    \centering
        \includegraphics[width=1\linewidth]{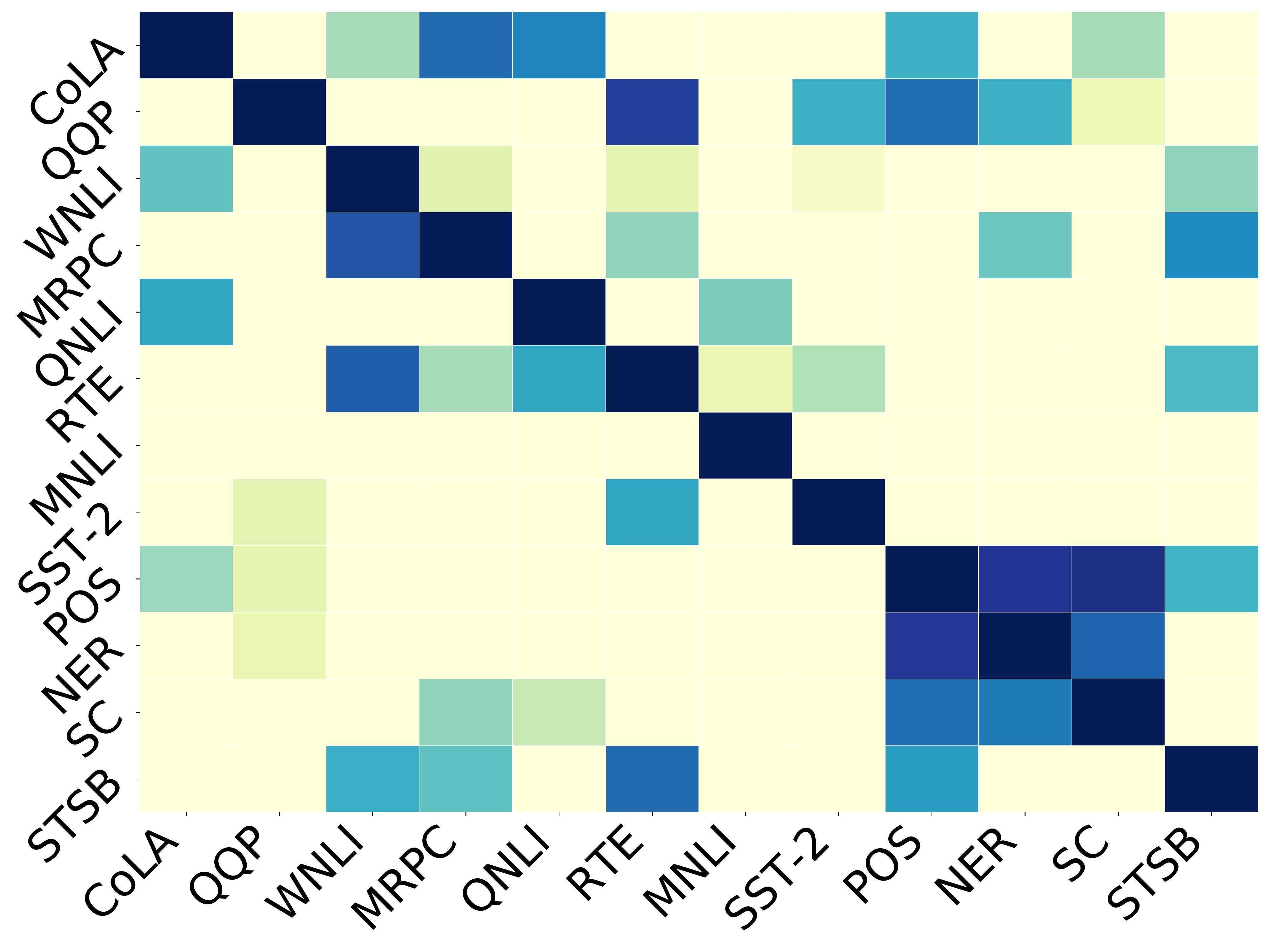}
        \caption{GradTS-fg}
    \end{subfigure}
    \caption{Task selection results by two GradTS methods on 12 tasks with mixed objectives. Y and X axes represent primary and auxiliary tasks, respectively.}
    \label{fig:task-ranking-multi-objective}
\end{figure}

As Table \ref{tbl:runtime} shows, the average GPU usage is comparable for all the four auxiliary task selection methods in the main experiments.
All the experiments are run with a batch size of 32 on an NVIDIA RTX-8000 graphics card.

Among the four methods, GradTS-trial is the most time-efficient mainly because its task rankings are generated from single-task experiments and they are fixed for all the evaluations.
While GradTS-fg filters training instances based on the output of GradTS-trial, the additional time cost is only linearly correlated with the training data size of auxiliary tasks.
On average, GradTS-fg takes longer time to finish than AUTOSEM-p1 but is more efficient than AUTOSEM.
Since GradTS reuses the task-specific head importance matrices and the thresholds for subsampling auxiliary tasks, it becomes gradually more time-economic than AUTOSEM and AUTOSEM-p1 when the candidate task set is larger or growing.
Thus, GradTS is a superior choice to AUTOSEM on large and complex task sets in terms of both efficacy and efficiency.

\section{Discussions} \label{sct:discussions}
GradTS is shown to be effective on 8 classification NLU tasks in our main experiments. 
In this section, we conduct case studies to (1) explore whether GradTS is effective on tasks that are difficult or have different training objectives, (2) validate that GradTS selects better auxiliary task sets than human intuition, and (3) justify our use of bert-base-cased as the backend model of GradTS and the MTL evaluation framework.

\begin{table*}[t]
\centering
\begin{tabular}{|l|r|r|r|r|r|r|}
\hline
\multicolumn{1}{|c|}{}                                                                           & \multicolumn{6}{c|}{Methods}                                                                                                                                                                          \\ \cline{2-7} 
\multicolumn{1}{|c|}{\multirow{-2}{*}{\begin{tabular}[c]{@{}c@{}}Primary \\ Tasks\end{tabular}}} & \multicolumn{1}{c|}{AUTOSEM-p1} & \multicolumn{1}{c|}{AUTOSEM} & \multicolumn{1}{c|}{GradTS-trial} & \multicolumn{1}{c|}{GradTS-fg} & \multicolumn{1}{c|}{Single-Task} & \multicolumn{1}{c|}{NO-SEL}  \\ \hline
CoLA                                                                                             & 57.04                           & 56.50                        & 60.08                             & \textbf{60.14}                 & 51.00                            & 47.52                        \\ \hline
MRPC                                                                                             & 79.66/85.21                     & 83.09/87.79                  & 83.58/88.35                       & \textbf{84.56/88.89}           & 78.19/84.58                      & 80.88/86.22                  \\ \hline
MNLI                                                                                             & 84.01/83.38                     & 83.80/83.65                   & \textbf{83.79/83.96}              & 84.05/83.67                    & 83.52/83.88                      & 83.10/82.95                  \\ \hline
QNLI                                                                                             & 90.88                           & 90.61                        & 90.50                             & \textbf{91.01}                 & 90.26                            & 89.73                        \\ \hline
QQP                                                                                              & 90.94/87.77                     & 90.61/87.54                  & 90.65/87.39                       & \textbf{90.96/87.81}           & 90.34/87.15                      & 90.30/86.82                  \\ \hline
RTE                                                                                              & 69.68                           & 70.04                        & 77.62                             & \textbf{79.42}                 & 63.18                            & 74.73                        \\ \hline
SST-2                                                                                            & 90.94                           & 91.17                        & 92.32                             & \textbf{92.81}                 & 91.63                            & 92.20                        \\ \hline
WNLI                                                                                             & 54.93                           & 52.11                        & 69.01                             & \textbf{71.97}                 & 53.52                            & 61.97 \\ \hline
STSB                                                                                             & 87.90/87.67                      & 89.12/88.79                  & 89.07/88.78                       & \textbf{89.26/88.90}           & 86.35/86.30                      & 86.63/86.80                  \\ \hline
POS                                                                                              & 91.43                           & 91.53                        & 91.60                              & \textbf{91.86}                 & 91.60                            & 90.42                        \\ \hline
NER                                                                                              & 91.23                           & 91.70                        & 92.55                             & \textbf{92.69}                 & 90.96                            & 88.80                         \\ \hline
SC                                                                                               & 90.22                           & 89.26                        & 93.39                             & \textbf{93.76}                 & 87.67                            & 87.99                        \\ \hline
\end{tabular}
\caption{MTL evaluation results with AUTOSEM and GradTS methods on 12 tasks with mixed training objectives. NO-SEL indicates performance of the MTL model trained on all 12 tasks.}
\label{tbl:eval-res-multi-objective}
\end{table*}
\subsection{Task Selection with Difficult Tasks}
GradTS relies on the hypothesis that the amount of gradients distributed on each attention head reflects the important linguistic features for a task.
However, tasks that are difficult for a model introduce more noise to its gradient calculations and thus may have negative effects on GradTS.
To study the effect of difficult tasks, we evaluate GradTS on a task set containing the 8 GLUE classification tasks and two MELD tasks.
The MELD and Dyadic-MELD tasks are difficult for the bert-base-cased model as the single-task performance on these tasks are both below 50 in F-1 scores.

We note that the largest tasks in size, i.e. MNLI and QQP, are not chosen as auxiliary tasks for either MELD or Dyadic-MELD, suggesting that training data amount is not a decisive factor for auxiliary task selection.
As auxiliary tasks, MELD is selected for SST-2 and CoLA, and Dyadic-MELD for SST-2 and RTE.
The connection between SST-2 and the two MELD tasks is intuitive since emotional and sentiment features are interconnected, while the other selections are not as intuitive.

We show the evaluation scores in Table \ref{tbl:res-cls-with-meld}.
Compared to Table \ref{tbl:main-experiment-results}, we note that 
the performance of AUTOSEM-p1 is largely harmed when MRPC, MNLI, and WNLI are set up as primary tasks, while AUTOSEM performance also suffers on QQP.
On the contrary, GradTS-trial performs relatively stably and GradTS-fg frequently produces auxiliary task sets higher in quality on the enlarged candidate task sets than on the 8 GLUE classification tasks only.
We attribute the strength of GradTS-fg to its ability to discard noisy training instances and mainly select datapoints contributing to the primary tasks.
When MELD and Dyadic-MELD are primary tasks, MTL performance, either with or without auxiliary task selection, is generally higher than the single-task baseline.
These results indicate the importance of MTL research and highlight the study of good auxiliary task selection methods, especially on tasks that are difficult under the single-task setting.
We also note that while AUTOSEM-p1 is not able to generate high-quality auxiliary task sets for MELD, the successive data subsampling mechanism in AUTOSEM polishes the data selection and improves the MTL performance by 2.63 in F-1 score.
Similarly, GradTS-fg generates better auxiliary task sets than GradTS-trial in all the evaluations, revealing the necessity of filtering out noisy auxiliary training instances.
To conclude, while both GradTS-trial and GradTS-fg are robust to difficult tasks in the candidate task sets, GradTS-fg is, in general, more optimal in these scenes.
\begin{table*}[t]
\centering
\begin{tabular}{|c|c|c|c|c|c|c|c|c|}
\hline
\multirow{2}{*}{Methods} & \multicolumn{8}{c|}{Primary Tasks}                                                                                                                      \\ \cline{2-9} 
                         & CoLA           & MRPC                 & MNLI                 & QNLI           & QQP                  & RTE            & SST-2          & WNLI           \\ \hline
Single-Task              & 51.00          & 78.19/84.58          & 83.52/83.88          & 90.26          & 90.34/87.15          & 63.18          & 91.63          & 53.52          \\ \hline
HEU-Size                 & 53.15          & 80.39/86.30          & 83.47/83.39          & 90.50          & 90.82/87.61          & 66.43          & 91.40          & 49.30          \\ \hline
HEU-Type                 & 54.44          & 80.15/86.66          & 83.52/83.32          & 90.61          & 90.71/87.50          & 73.65          & 91.86          & 54.92          \\ \hline
HEU-Len                  & 54.17          & 80.64/85.82          & 83.36/83.34          & 90.39          & 90.57/87.24          & 67.50          & 91.63          & 52.11          \\ \hline
GradTS-trial                                   & 55.24          & 83.58/88.35          & 83.95/83.55          & 90.62          & 90.87/87.72          & 76.53          & 92.31          & 54.93          \\ \hline
GradTS-fg                                      & \textbf{58.38} & \textbf{84.07/88.74} & \textbf{83.79/83.96} & \textbf{90.87} & \textbf{90.89/87.73} & \textbf{76.90} & \textbf{92.63} & \textbf{57.75} \\ \hline
\end{tabular}
\caption{MTL evaluation results on 8 GLUE classification tasks. HEU-Size, HEU-Type, and HEU-Len refer to MTL performance with intuitive auxiliary task selections based on training data size, task type, and average sentence length, respectively.}
\label{tbl:intuition-discussion}
\end{table*}
\subsection{Task Selection with Mixed Objectives}
Including AUTOSEM, most prior publications on MTL consider only auxiliary tasks with the same training objective as the primary task.
This overly simplifies the auxiliary task selection problem and limits the scope of research on the topic.
In this section, we examine the applicability of GradTS to candidate task sets with mixed objectives.
As candidate tasks, we use the 8 GLUE classification tasks, a regression task (STSB), and three sequence labeling tasks (POS, NER, and SC).
The auxiliary task selection results by GradTS are shown in Figure \ref{fig:task-ranking-multi-objective}, which make intuitive sense in some cases, e.g. POS and SC are closely bond to CoLA and STSB is selected as an auxiliary task for MRPC.

We assess the quality of auxiliary task sets produced by GradTS via evaluations with MT-DNN and display the results in Table \ref{tbl:eval-res-multi-objective}.
Results show that the performance of GradTS does not suffer from introducing the four non-classification tasks, as the auxiliary task sets selected by GradTS in most cases lead to higher MTL performance than in Table \ref{tbl:main-experiment-results}.
In comparison, the auxiliary task sets produced by AUTOSEM-p1 and AUTOSEM are noisier with the four newly-introduced tasks, causing noticeable performance drops to 3 and 2 GLUE classification tasks, respectively.
Furthermore, while both GradTS-trial and GradTS-fg lead to higher MTL performance than not applying any auxiliary task selection method, applying AUTOSEM-p1 and AUTOSEM causes performance drops in 4 and 3 tasks, respectively.
AUTOSEM-p1 and AUTOSEM even cause the MTL performance to drop below the single-task evaluation scores in 3 and 4 experiments, respectively.
The results indicate that, despite the potentially increased discrepancies among tasks with various objectives, GradTS is an effective and robust auxiliary task selection method.
We also note that since GradTS reuses the head ranking matrices produced in the main experiments, its additional time cost on the enlarged task set is negligible, compared to AUTOSEM which has to be fully re-run.
This further demonstrates the efficiency of GradTS, especially when the candidate task set grows larger.

\subsection{Comparison to Intuitive Task Selections}
\begin{table}[t]
\centering
\begin{tabular}{|c|c|c|c|}
\hline
Datasets & SIZE & LEN & TYPE          \\ \hline
CoLA     & 8,550         &7.70& Single-Sentence \\ \hline
MRPC     & 3,667         &21.77& Paraphrase \\ \hline
MNLI     & 392,701       &15.34& Inference \\ \hline
QNLI     & 104,742       &18.22& Inference    \\ \hline
QQP      & 363,845       &11.06&Paraphrase\\ \hline
RTE      & 2,489         &26.18&Inference\\ \hline
SST-2    & 67,348        &9.41&Single-Sentence\\ \hline
WNLI     & 634           &13.90&Inference\\ \hline
\end{tabular}
\caption{Specifics of 8 GLUE classification tasks. SIZE, LEN, and TYPE indicate the number of training instances, average sentence length in terms of words, and task type, respectively. The task types are as defined in \citet{glue-orig}.}
\label{tbl:task-specs-intuition}
\end{table}
\begin{table*}[th!]
\centering
\begin{tabular}{|l|r|r|r|r|r|r|}
\hline
\multicolumn{1}{|c|}{\multirow{2}{*}{Methods}} & \multicolumn{6}{c|}{Primary Tasks}                                                                                                                                    \\ \cline{2-7} 
\multicolumn{1}{|c|}{}                         & \multicolumn{1}{c|}{CoLA} & \multicolumn{1}{c|}{MRPC} & \multicolumn{1}{c|}{QNLI} & \multicolumn{1}{c|}{RTE} & \multicolumn{1}{c|}{SST-2} & \multicolumn{1}{c|}{WNLI} \\ \hline
bert-base-uncased                              & 51.15                     & 87.00/81.61               & 90.97                     & 68.95                    & 91.85                      & 50.70                     \\
+GradTS-trial                                  & 55.99                     & 87.49/82.84               & 90.86                     & 69.31                    & 91.97                      & 60.57                     \\
+GradTS-fg                                     & 56.16                     & 88.38/83.82               & 91.25                     & 71.11                    & 92.08                      & 69.01                     \\ \hline
bert-base-cased                                & 52.59                     & 87.87/83.08               & 90.02                     & 67.14                    & 91.28                      & 53.52                     \\
+GradTS-trial                                  & 56.49                     & 88.14/83.57               & 90.31                     & 68.95                    & 91.97                      & 64.79                     \\
+GradTS-fg                                     & 59.55                     & 88.65/84.06               & 90.60                     & 71.48                    & 92.20                      & 74.65                     \\ \hline
bert-large-uncased                             & 57.52                     & 81.62/86.58               & 91.10                     & 67.87                    & 92.55                      & 42.25                     \\
+GradTS-trial                                  & 56.06                     & 82.11/87.89               & 91.67                     & 68.95                    & 92.20                      & 66.19                     \\
+GradTS-fg                                     & 58.04                     & 82.60/87.95               & 91.74                     & 71.84                    & 93.35                      & 76.06                     \\ \hline
bert-large-cased                               & 57.60                     & 82.84/87.72               & 91.91                     & 66.79                    & 92.67                      & 63.38                     \\
+GradTS-trial                                  & 59.35                     & 83.82/88.30               & 91.98                     & 71.12                    & 93.12                      & 73.24                     \\
+GradTS-fg                                     & 60.33                     & 85.53/89.56               & 92.37                     & 74.37                    & 93.92                      & 78.87                     \\ \hline
roberta-base                                   & 51.26                     & 87.01/91.39               & 92.11                     & 68.59                    & 93.46                      & 50.71                     \\
+GradTS-trial                                  & 55.99                     & 87.26/90.91               & 92.57                     & 70.40                    & 93.81                      & 64.78                     \\
+GradTS-fg                                     & 56.00                     & 87.99/91.39               & 92.73                     & 71.12                    & 94.15                      & 66.19                     \\ \hline
roberta-large                                  & 59.57                     & 79.90/86.82               & 92.17                     & 74.73                    & 95.41                      & 52.11                     \\
+GradTS-trial                                  & 63.68                     & 85.54/89.70               & 93.80                     & 79.42                    & 95.64                      & 56.34                     \\
+GradTS-fg                                     & 64.56                     & 86.77/90.79               & 94.14                     & 80.51                    & 96.22                      & 60.56                     \\ \hline
\end{tabular}
\caption{MTL evaluation results with and without task selection methods on 6 GLUE tasks. Rows with the model names indicate the multi-task evaluation results of each model on the entire candidate task set.}
\label{tbl:eval-results-model-selection}
\end{table*}
We further validate the strength of GradTS by comparing the MTL performance with GradTS to that with three intuitive task selection methods based on simple dataset analysis.
The three heuristics we set up for comparison choose auxiliary tasks based on (1) training data size; similarity between the primary and auxiliary tasks with respect to (2) task type and (3) average sentence length.
Table \ref{tbl:task-specs-intuition} displays the training data amount, task type, and average sentence length of the 8 tasks.
For HEU-Size and HEU-Len, starting from the most appropriate auxiliary task, we keep adding tasks into the auxiliary task set greedily and report the best score.

According to Table \ref{tbl:intuition-discussion}, while the intuitive task selections usually result in higher performance than the single-task evaluation scores (and comparable to AUTOSEM results shown in Table \ref{tbl:main-experiment-results}), the GradTS methods outperform the intuitive methods on all the 8 tasks.
Among the three intuitive task selection methods, HEU-Type in most cases produces the auxiliary task sets highest in quality.
This demonstrates the high priority of auxiliary tasks with similar goals as the primary task at the task selection phase.
While the importance of task types is reflected in the task selection results of GradTS (Figure \ref{fig:aux-task-rankings}) as well, GradTS is able to take other empirical clues into consideration and construct more effective auxiliary task sets.
These additional clues, however, are expensive to design and cannot be directly transferred to other candidate task sets without costly adaptations if manual auxiliary task selection methods are applied.
Moreover, the three simple heuristics are not always applicable when the candidate task set becomes complex, e.g. containing tasks with varying label sets or with multiple objectives.
GradTS, on the contrary, has shown great capability and robustness in these complex cases with moderate time and resource cost.
It is a promising method in place of expensive manual auxiliary task set design in MTL research.

\section{Base Model Selection for GradTS} \label{sct:model-selection-discussion}
Since GradTS is built on Transformer-based models, we select its backend model from 6 common pre-trained Transformer-based models, namely bert-base-uncased, bert-base-cased, bert-large-uncased, bert-large-cased, roberta-base, and roberta-large.
We set up MTL evaluations with MT-DNN on CoLA, MRPC, SST-2, WNLI, QNLI, and RTE to examine the appropriateness of these backend models in GradTS.
Specifically, we assess the strength and robustness of these models by comparing the MT-DNN performance with auxiliary task sets selected by GradTS against that without auxiliary task filtering.
The same backend model is used for GradTS and MT-DNN in each experiment to eliminate possible discrepancies across models.

From Table \ref{tbl:eval-results-model-selection}, we notice clear performance gaps between the cased and uncased models as the backend of GradTS.
For example, on CoLA and SST-2, GradTS-trial produces worse auxiliary task sets than using all candidate tasks with a bert-base-uncased backend, while GradTS with a bert-base-cased backend improves model performance for both GradTS-trial and GradTS-fg.
This is intuitive since case information is crucial for grammaticality and sentiment tasks.
Among the four cased backend models, RoBERTa \cite{roberta-orig} does not trigger larger MT-DNN performance improvements than BERT of the same size, implying that larger pre-training corpora do not greatly affect the efficacy of GradTS.
While the performance improvement brought by GradTS with the bert-large-cased backend is comparable to that with the bert-base-cased backend, its running time and GPU cost are over 100\% higher.
We thus choose bert-base-cased as the backend of GradTS to balance performance with resource cost, though potentially any cased Transformer-based model is a valid choice.

\section{Conclusion and Future Work}
This paper presented GradTS, an automatic auxiliary task selection method for MTL based on pre-trained Transformer-based models.
On 8 GLUE classification tasks, GradTS produced auxiliary task sets higher in quality than AUTOSEM, a strong baseline method, with less time and resource consumption.
In our case studies comparing GradTS to intuitive task selections,
GradTS showed greater capability of finding more optimal auxiliary task sets than using trivial heuristics based on statistics of datasets.
We additionally demonstrated that GradTS was both more effective and more efficient than our baseline on
task sets with mixed objectives.
GradTS was also shown to be robust in task sets containing difficult tasks for its backend model.
These findings support the applicability of GradTS to a wide range of task and model settings.
Future work may extend the use of GradTS to determine high-quality auxliary datasets for the same task.


\bibliography{tacl2018}
\bibliographystyle{acl_natbib}

\clearpage
\newpage

\appendix
\section{Head and Task Ranking Examples}
\begin{figure}[h]
    \centering
    \includegraphics[height=.55\linewidth]{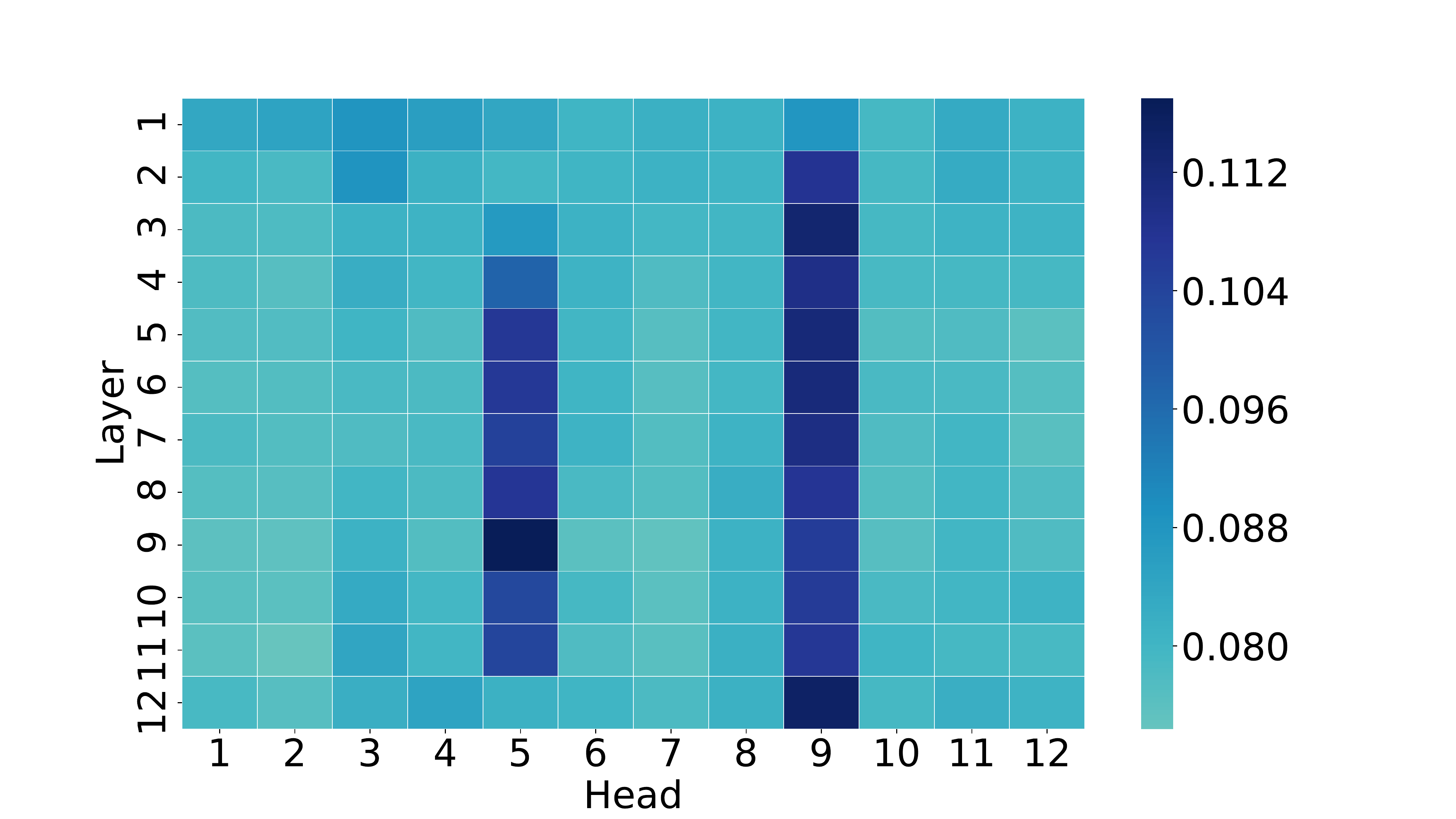}
    \caption{Example head importance matrix of bert-base-cased on MRPC. Darker color indicate higher importance.}
    \label{fig:head-ranking-example}
\end{figure}
\begin{figure}[h]
    \centering
    \includegraphics[height=.55\linewidth]{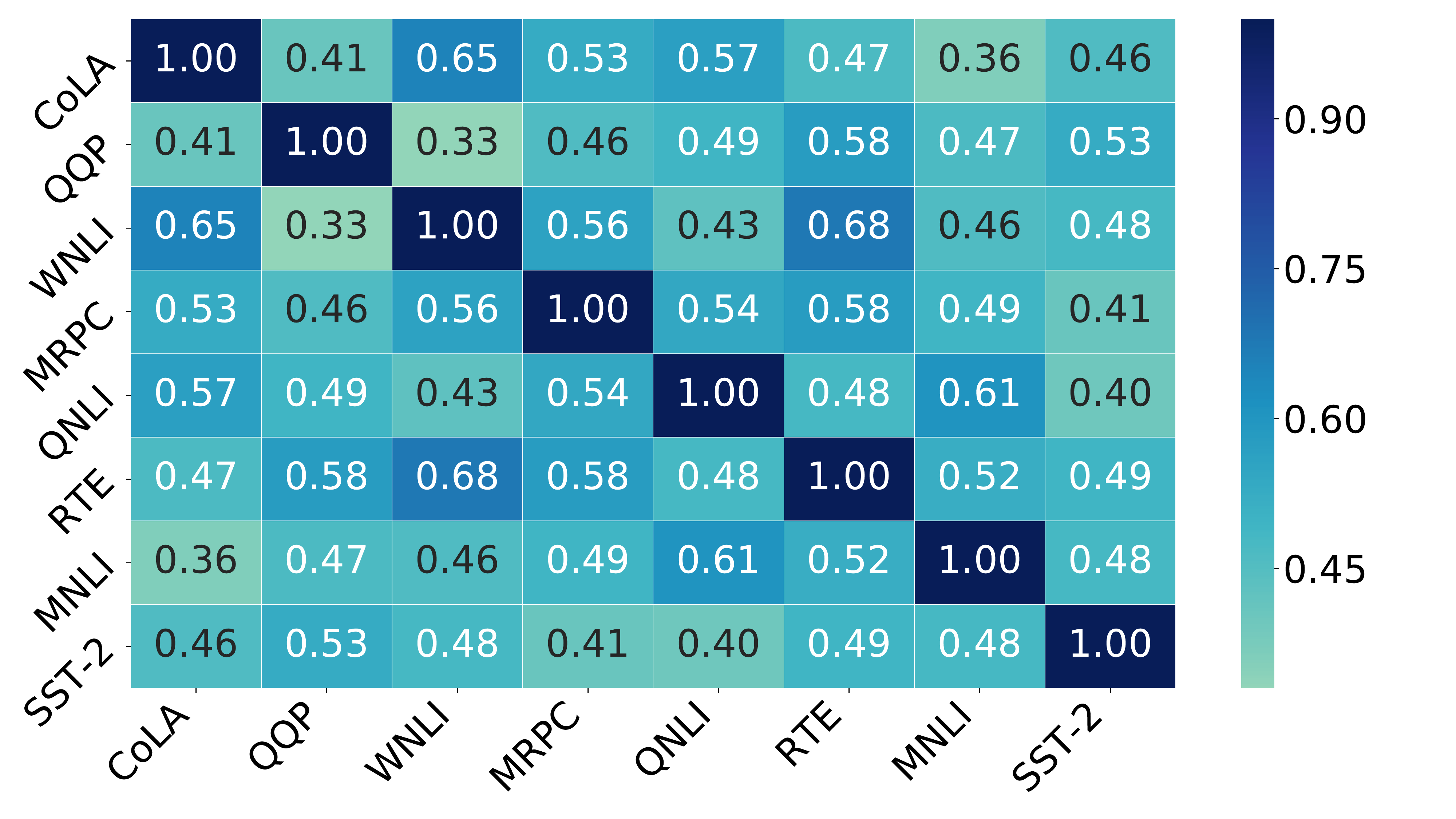}
    \caption{Kendall's rank correlations among 8 GLUE classification tasks generated by the auxiliary task ranking module of GradTS.}
    \label{fig:task-corr}
\end{figure}
We show the head importance matrix of the bert-base-cased model on MRPC in Figure \ref{fig:head-ranking-example}.
The task correlation matrix generated by bert-base-cased on the 8 GLUE classification tasks is shown in Figure \ref{fig:task-corr}.
\section{Threshold-based Task Selection}
\begin{table}[ht]
\small
\centering
\begin{tabular}{|c|c|c|c|}
\hline
Dataset & GradTS-thres & GradTS-trial & GradTS-fg            \\ \hline
CoLA                                                   & 54.69        & 55.24        & \textbf{58.38}       \\ \hline
MRPC                                                   & 83.20/83.59  & 83.58/88.35  & \textbf{84.07/88.74} \\ \hline
MNLI                                                   & 83.81/83.79  & 83.95/83.55  & \textbf{83.79/83.96} \\ \hline
QNLI                                                   & 90.61        & 90.62        & \textbf{90.87}       \\ \hline
QQP                                                    & 90.47/87.28  & 90.87/87.72  & \textbf{90.89/87.73} \\ \hline
RTE                                                    & 66.98        & 76.53        & \textbf{76.90}       \\ \hline
SST-2                                                  & 91.63        & 92.31        & \textbf{92.63}       \\ \hline
WNLI                                                   & 56.34        & 54.93        & \textbf{57.75}       \\ \hline
\end{tabular}
\caption{MTL evaluation results of the three settings of GradTS on 8 GLUE classification tasks. The highest score for each task is in bold.}
\label{tbl:eval-results-thres}
\end{table}
\begin{table}[]
\centering
\begin{tabular}{|l|r|r|}
\hline
\multicolumn{1}{|c|}{Methods} & \multicolumn{1}{c|}{Time Cost} & \multicolumn{1}{c|}{GPU Usage} \\ \hline
AUTOSEM-p1                    &114&37,003\\ \hline
AUTOSEM                       &194&46,361\\ \hline
GradTS-thres                  &87&27,481\\ \hline
GradTS-trial                  &107&39,551\\ \hline
GradTS-fg                     &1547&35,610\\ \hline
\end{tabular}
\caption{Average time and GPU consumption for 5 auxiliary task selection methods on each of the 8 GLUE classification tasks. The units are minutes for time cost and megabytes for GPU usage.}
\label{tbl:cost-thres}
\end{table}
Besides the two settings of GradTS (GradTS-trial and GradTS-fg) for the auxiliary task selection module, we additionally introduce a task-level threshold-based setting.
Under this setting, we empirically choose a threshold with which GradTS produces the best auxiliary task sets in a small collection of tasks.
Then for each primary task, GradTS selects all the tasks having correlation scores above the threshold into the auxiliary task set.
We use Kendall's $\tau$ of 0.47 as the threshold in the evaluations, which is tuned via experiments on RTE, MRPC, and CoLA tasks.
We refer to GradTS with the task-level threshold-based setting as GradTS-thres.

It is worth noting that though GradTS-thres is more time- and resource-economic than GradTS-trial and GradTS-fg, its performance is the lowest in most cases.
We display the evaluation results of GradTS-thres, GradTS-trial, and GradTS-fg on the 8 GLUE classification tasks in Table \ref{tbl:eval-results-thres} and the average time and memory cost in Table \ref{tbl:cost-thres}.
Since GradTS-thres is not optimal compared to GradTS-trial and GradTS-fg, and sometimes outperformed by AUTOSEM, we do not introduce this setting in the main body of our paper.
However, it is still an interesting method to study regarding its time- and resource-efficiency and stronger capability than not applying any auxiliary task selection method.
\end{document}